\pdfoutput=1 
\documentclass{article}
\usepackage[preprint]{spconf}
\usepackage{amsmath,graphicx,booktabs,microtype}
\copyrightnotice{\parbox[t]{\textwidth}{\centering\footnotesize This work has been submitted to the
  IEEE for possible publication.\\ Copyright may be transferred without notice, after which this
  version may no longer be accessible.}}
\graphicspath{{figures/}}
\usepackage[hidelinks]{hyperref}
\hypersetup{pdftitle={Found but Not Read: When Extracted Text Closes the Retrieval--Reading Gap in Document
  Vision--Language Models},
  pdfauthor={Qingtao Xia, Siyao Cheng, Jiahua Bao, Jiaxing Du, and Jie Liu}}
\ninept 

\makeatletter
\long\def\@makecaption#1#2{%
 \vskip 4pt
 \setbox\@tempboxa\hbox{#1. #2}%
 \ifdim \wd\@tempboxa >\hsize #1. #2\par \else \hbox to\hsize{\hfil\box\@tempboxa\hfil}\fi}
\let\spconf@thebibliography\thebibliography
\def\thebibliography#1{\spconf@thebibliography{#1}\setlength{\itemsep}{1.5pt plus 1pt}}
\makeatother

\newcommand{\dt}{\Delta}

\title{Found but Not Read: When Extracted Text Closes the\\Retrieval--Reading Gap in Document Vision--Language Models}

\name{Qingtao Xia, Siyao Cheng\sthanks{Corresponding author.},
Jiahua Bao, Jiaxing Du, and Jie Liu}
\address{Harbin Institute of Technology, Harbin, China}

\begin{document}
\maketitle

\begin{abstract}
Retrieval-augmented document question answering assumes that once the right page is found, a
vision--language model (VLM) can read it. We show that this assumption often fails, leaving a
retrieval--reading gap: evidence found but not used. A paired protocol isolates this gap by
comparing answers from the retrieved page images alone with answers from the same images plus
their extracted text. On FoveDoc-Bench, our benchmark with traceable evidence, retrieval finds
nearly every evidence page, yet adding CPU-OCR text raises strict accuracy by $13$ to $16$
points. An exact text layer roughly doubles the gain, which appears across six VLMs from three
families and, within one family, narrows with scale without closing. The reader can read this
evidence but cannot find it: crops of it recover most of the text gain, boxes around it on the
page none. The same protocol identifies two boundaries. Extracted text helps on textual evidence
but is neutral or harmful on charts and figures. Its advantage shrinks as retrieval degrades, and
unrelated text of the same form adds nothing detectable. Extracted text is an amplifier of
retrieval that works, not a substitute for retrieval that does not. Our code is available at
\url{https://github.com/atoz03/fovedoc-sup}.
\end{abstract}


\section{Introduction}
\label{sec:intro}

Answering a question about a long document takes two steps: find the pages that hold the
evidence, then read them. A retrieval-augmented pipeline~\cite{rag} gives the first step to a
visual retriever~\cite{colpali,visrag,m3docrag} and the second to a VLM that receives the
top-ranked page images. Progress is tracked by answer accuracy~\cite{docvqa,mpdocvqa,mmlb} and
retrieval recall, and the premise that links them is rarely stated: once the right page is in the
context window, reading it is the easy part. Neither number tests the premise, because neither
separates a question the retriever lost from one the reader lost.

\begin{figure*}[t]
\centering
\includegraphics[width=.96\textwidth]{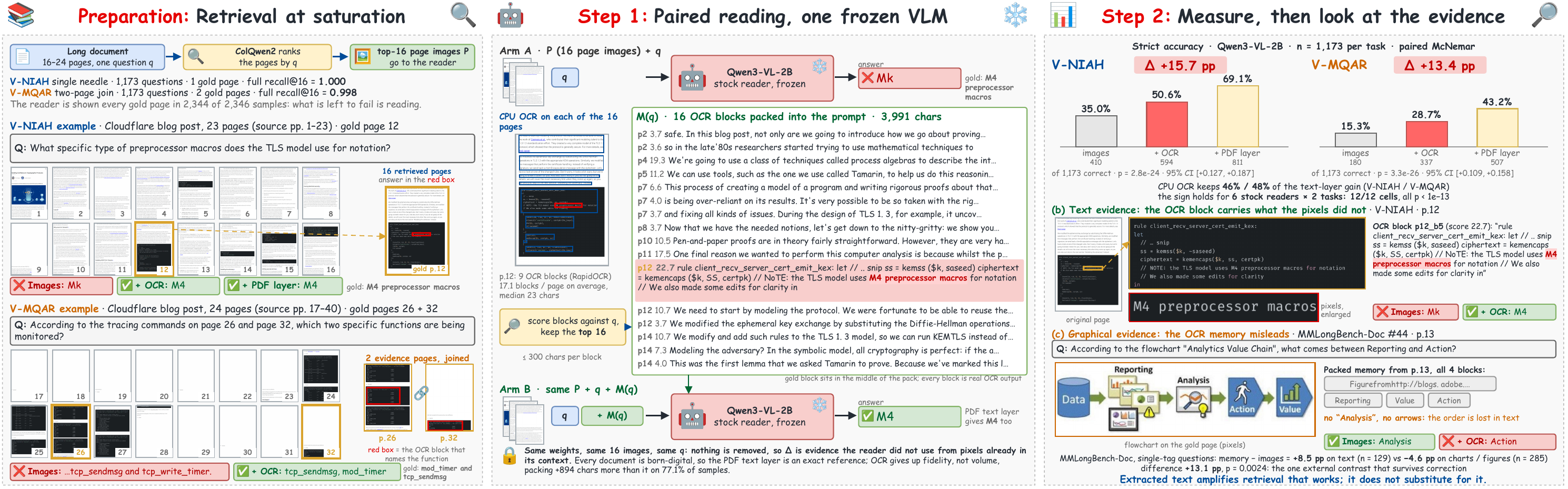}
\caption{The paired protocol and its headline result. \textbf{Preparation:} ColQwen2 retrieves
$16$ page images per question at saturated recall; one V-NIAH and one V-MQAR sample are shown
verbatim. \textbf{Step 1:} one frozen VLM answers from the images (arm A) and from the same
images plus the $16$ CPU-OCR blocks scoring highest against the question (arm B), so $\dt$ is
evidence the reader did not use from pixels already in its context. \textbf{Step 2:} strict
accuracy (Qwen3-VL-2B, $n{=}1173$ per task; pp $=$ percentage points) and one verbatim evidence
case per direction.}
\label{fig:design}
\end{figure*}

Our protocol separates them. A page can reach the reader as a pixel array, as in OCR-free
models~\cite{donut,pix2struct}, or as a symbolic transcript, as in OCR-augmented
ones~\cite{docvlm}, and a VLM may use evidence in the second form that it fails to extract from
the first~\cite{ocrbench,pixelworld}. The protocol (Fig.~\ref{fig:design}) holds retrieval at
saturation, so nearly every evidence page is in front of the reader, and asks the same question
twice: once from the retrieved page images, once from the same images plus a text memory
extracted from them. Nothing is removed, so any improvement is evidence found but not used, the
\emph{retrieval--reading gap}. Text from a CPU OCR engine, the only text a scanned document can
offer, adds $13$ to $16$ points of strict accuracy; the exact born-digital text layer roughly
doubles that. The gap is not a quirk of one backbone: with the text layer it appears in all six
stock readers we test from three model families, and within one family it narrows from 2B to 8B
parameters without closing. Reading is not the easy part.

The reader can read this evidence; what it cannot do is find it. Crops of the same evidence
blocks, with no text at all, recover $87\%$ and $92\%$ of the text gain, while boxes drawn
around them on the full pages recover none. Page retrieval leaves a second localization problem
inside the context window.

The obvious lesson, OCR every page, is the wrong one, and the same protocol identifies two
boundaries of the benefit. Extracted text helps where the evidence is text and is neutral to
harmful where it is a chart or figure; on MMLongBench-Doc~\cite{mmlb}, whose evidence is mostly
graphical, the two effects cancel. The advantage also shrinks as retrieval degrades: when we
withdraw gold pages from the retrieved set under a fixed page budget, most of the advantage
leaves with them, the rest is the document's own context, and unrelated text of the same form
adds nothing detectable. The text does not help because text is easier to consume than pixels;
it helps because it delivers evidence the retriever found. Extracted text is an amplifier of
retrieval that works, not a substitute for retrieval that does not.

These interventions need evidence that can be traced, withdrawn and re-presented, so we build
\textbf{FoveDoc-Bench}: $2346$ questions over $1173$ born-digital documents, each annotated with
its evidence pages, blocks and excerpts. We contribute:

\begin{enumerate}\itemsep0pt\parskip0pt
\item a \textbf{retrieval-controlled paired protocol} that isolates the retrieval--reading gap:
      $13$--$16$ points with CPU OCR, roughly twice that with an exact text layer, which holds
      in six readers (Secs.~\ref{sec:setup}--\ref{sec:gap});
\item an \textbf{evidence-accessibility account}: crops recover $87$--$92\%$ of the text gain
      and boxes none, and evidence modality and retrieval quality bound the benefit
      (Secs.~\ref{sec:gap}--\ref{sec:recall});
\item \textbf{FoveDoc-Bench}, traceable evidence for every question, which makes such
      interventions possible (Sec.~\ref{sec:setup}).
\end{enumerate}

\section{The paired protocol and FoveDoc-Bench}
\label{sec:setup}

\begin{table}[t]
\centering
\footnotesize
\setlength{\tabcolsep}{3.5pt}
\caption{Benchmarks. Recall@16 is the fraction of questions whose every gold page is among the
$16$ ColQwen2 retrieves: FoveDoc-Bench is saturated, MMLongBench-Doc is not.}
\label{tab:bench}
\begin{tabular}{lcccc}
\toprule
benchmark & questions & pages/doc & gold pages & recall@16 \\
\midrule
V-NIAH & 1173 & 16--24 & 1 & 1.000 \\
V-MQAR & 1173 & 16--24 & 2 & 0.998 \\
MMLongBench-Doc & $1091^{\ast}$ & $\leq 468$ & $0$--$n$ & 0.844 \\
\bottomrule
\end{tabular}
\\[1pt]
{\scriptsize $^{\ast}$ $135$ documents; $838$ answerable with annotated evidence, $244$ unanswerable.}
\end{table}

\noindent\textbf{Arms.} ColQwen2~\cite{colpali} ranks the pages and the top $k{=}16$ go to the
reader, stock Qwen3-VL-2B-Instruct~\cite{qwen3vl} unless stated. The \textbf{images} arm
receives the $16$ pages as rendered images. The \textbf{memory} arm receives the same $16$
images plus a text memory: the $16$ text blocks from those pages that score highest lexically
against the question, grouped by page and tagged with page numbers, each capped at $300$
characters, so every memory is compared at one block budget. The \emph{OCR memory} is recovered
by RapidOCR (PP-OCR models~\cite{ppocr}) on CPU from the rendered page images; the \emph{PDF
text layer} is the exact born-digital text of the same pages. OCR errors propagate into
RAG~\cite{ohrbench}, so OCR is the headline and the text layer the upper bound. Both memories
come from pages the image arm already receives, so any gain is evidence the reader held and did
not use.

\noindent\textbf{FoveDoc-Bench.} We build FoveDoc-Bench from $1173$ born-digital documents, each
supplying a $16$--$24$-page context, numbered as in the source, and one question per task
(Table~\ref{tab:bench}).
\textbf{V-NIAH} asks for one evidence block on one page; \textbf{V-MQAR} joins two blocks on
distinct pages. Questions and answers are generated from source excerpts, and construction
checks that every answer is nonempty and every excerpt resolves to a source block, so each
question carries its answer, its page images, and the evidence pages, blocks and excerpts behind
it. All $1173$ documents are held out from the auxiliary training and validation QA pools. External evaluation
uses \textbf{MMLongBench-Doc}~\cite{mmlb}, whose questions carry the benchmark's own
evidence-modality tags and include deliberately unanswerable items.

\begin{table}[t]
\centering
\footnotesize
\setlength{\tabcolsep}{3.2pt}
\caption{\textbf{The retrieval--reading gap.} Strict accuracy, Qwen3-VL-2B, $n{=}1173$ per
task, paired on identical pages. Both memory arms add packed text to the $16$ page images.
``Retained'' is the OCR gain as a share of the text-layer gain. OCR contrasts: $95\%$ CIs
$[+0.109,+0.158]$ and $[+0.127,+0.187]$, both $p<10^{-23}$.}
\label{tab:parity}
\begin{tabular}{lcccccc}
\toprule
task & images & +OCR & +PDF & $\dt_{\text{OCR}}$ & $\dt_{\text{PDF}}$ & retained \\
\midrule
V-MQAR & 0.153 & \textbf{0.287} & 0.432 & $+0.134$ & $+0.279$ & 48\% \\
V-NIAH & 0.350 & \textbf{0.506} & 0.691 & $+0.157$ & $+0.342$ & 46\% \\
\bottomrule
\end{tabular}
\end{table}

\noindent\textbf{Scoring.} Answers are graded by an LLM judge~\cite{llmjudge}; \emph{strict}
accuracy counts only verdicts of \texttt{correct}. Main and external comparisons use GPT-5.4;
localization and recall controls use GPT-5.6 Luna. The judges agree on $98.8\%$ of strict
verdicts over $1963$ identical predictions, and all arms of a contrast are graded by one judge
in one session. Two authors also double-annotated $200$ samples by hand, drawn from the
discordant pairs between the image and OCR arms and including answers that differ from the
gold only in wording; their verdicts agree with the judge's on $96\%$ of them. Each contrast is estimated as $\hat{\dt}=(b-c)/n$ from the discordant pairs, $b$
samples the memory arm alone answers correctly and $c$ the image arm alone, and tested by
McNemar's test~\cite{mcnemar} on $(b,c)$ with continuity correction, under Bonferroni thresholds
where several contrasts are reported.

\section{The retrieval--reading gap}
\label{sec:gap}

\begin{figure}[t]
\centering
\includegraphics[width=\columnwidth]{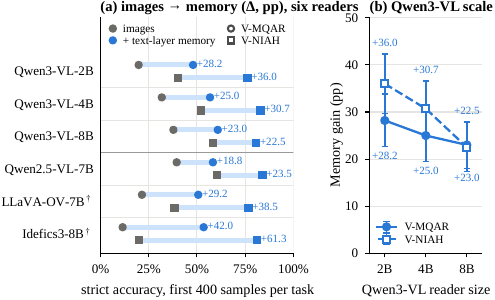}
\caption{Text-layer gains for six VLMs from three families. \textbf{(a)} Strict accuracy
from images (grey) to images plus text-layer memory (blue) for six stock readers on V-MQAR
(circles) and V-NIAH (squares), first $400$ samples per task; labels give the paired gain in pp,
every cell positive at $p<10^{-13}$. $^{\dagger}$Image splitting disabled or pixels capped to fit
$16$ pages. \textbf{(b)} Within Qwen3-VL at identical settings the gap narrows monotonically with
scale and is still $+0.23$ at 8B.}
\label{fig:readers}
\end{figure}

On FoveDoc-Bench full recall@16 is $1.000$ and $0.998$ (Table~\ref{tab:bench}): the reader is
shown every gold page in $2344$ of $2346$ questions, and what remains is reading.
Table~\ref{tab:parity} measures it for the two memories. CPU OCR of the retrieved pages, on top
of the same pages as images, raises strict accuracy by $+0.134$ on V-MQAR and $+0.157$ on V-NIAH;
the exact text layer roughly doubles both, and the ratio $\dt_{\text{OCR}}/\dt_{\text{PDF}}$ is
close to one half on both tasks. Relative to
the image arm the multi-hop task gains most: OCR memory nearly doubles V-MQAR accuracy ($0.153$
to $0.287$) and adds under half on the single needle ($0.350$ to $0.506$). What OCR gives up is
fidelity, not volume. At the fixed block budget the OCR arm packs \emph{more} characters than
the text-layer arm on $77\%$ of samples, $+894$ on average, because the born-digital layer
fragments into $53$ blocks per page at a six-character median where OCR line grouping yields
$17$ blocks at a median of $23$. What the OCR arm lacks is the evidence itself. Measured against
the text layer on FoveDoc-Bench's excerpts, the best single packable OCR block recovers $76\%$ of
a gold excerpt's tokens on average, and at least $90\%$ of them for $53\%$ of evidence items on
both tasks: the share of the text-layer gain that OCR retains tracks the share of gold evidence
that OCR returns intact.

\noindent\textbf{The gap is not a property of one backbone.} Fig.~\ref{fig:readers} repeats the
comparison for six stock readers across three families with the text-layer memory:
Qwen3-VL-2B/4B/8B~\cite{qwen3vl}, Qwen2.5-VL-7B~\cite{qwen25vl},
LLaVA-OneVision-7B~\cite{llavaov} and Idefics3-8B~\cite{idefics3}. The memory arm wins all twelve
(reader, task) cells at $p<10^{-13}$. At $512^2$ pixels a page costs $267$, $346$, $3036$ and
$3708$ image tokens for Qwen3-VL, Qwen2.5-VL, Idefics3 and LLaVA-OV, so the last two read $16$
pages with image splitting disabled or pixels capped and gain $+0.29$ to $+0.61$ there; the
Qwen readers, at native settings, gain $+0.19$ to $+0.36$, so across families only the sign is
comparable. Within Qwen3-VL, at identical settings, the gap narrows monotonically with scale,
from $+0.28$ to $+0.23$ on V-MQAR and from $+0.36$ to $+0.23$ on V-NIAH between 2B and 8B: a
stronger visual reader needs the text less, and at 8B still needs it.

\noindent\textbf{Cheaper levers do not close it.} With a 4B reader on V-MQAR, doubling the block
budget, a union selector and a reasoning-mode reader each move accuracy by at most $+0.03$;
self-ask decomposition~\cite{selfask} costs $-0.043$; a 2B LoRA~\cite{lora} recovers $+0.06$ on
V-MQAR and $+0.05$ on V-NIAH, and a 4B LoRA adds $+0.009$. Every lever is small next to the gap.

\noindent\textbf{The reader can read the evidence but cannot find it.} Three paired controls
locate the failure (V-MQAR / V-NIAH, $1173$ samples each). The full OCR text of the $16$ pages in
reading order, with no question-conditioned selection, gains $+0.179$ / $+0.221$, more than the
$16$-block digest under a neutral prompt ($+0.141$ / $+0.165$). The same blocks passed as image
crops at the reader's own page pixel density, appended to the pages without any text, gain
$+0.123$ / $+0.152$, which is $87\%$ / $92\%$ of the digest's gain. Red boxes around the same
blocks on the page images gain $+0.005$ / $-0.030$. The crops carry pixels the pages already
held: the reader can read the evidence but cannot find it among $16$ dense pages, and lifting it
out, as text or as a crop, makes it usable. Page-level retrieval is not evidence-level retrieval.

\noindent\textbf{Takeaway.} With every gold page in view, the reader still leaves $13$ to $16$
points of evidence unused in every backbone we test; lifting it off the page recovers it, with
or without question-conditioned selection.

\section{The modality boundary}
\label{sec:boundary}

\begin{table}[t]
\centering
\footnotesize
\setlength{\tabcolsep}{3.5pt}
\caption{MMLongBench-Doc, paired, strict accuracy, same reader, retriever, memory builder and
judge as Table~\ref{tab:parity}; Bonferroni $\alpha{=}0.0125$ over the four strata.}
\label{tab:external}
\begin{tabular}{lrcccc}
\toprule
stratum & $n$ & images & +mem. & $\dt$ & $p$ \\
\midrule
answerable, annotated & 838 & 0.220 & 0.223 & $+0.004$ & 0.86 \\
\ \ full recall@16    & 706 & 0.227 & 0.242 & $+0.016$ & 0.32 \\
\ \ partial recall    & 132 & 0.182 & 0.121 & $-0.061$ & 0.099 \\
unanswerable          & 244 & 0.041 & 0.090 & $+0.049$ & 0.014 \\
\bottomrule
\end{tabular}
\end{table}

\begin{figure}[t]
\centering
\includegraphics[width=\columnwidth]{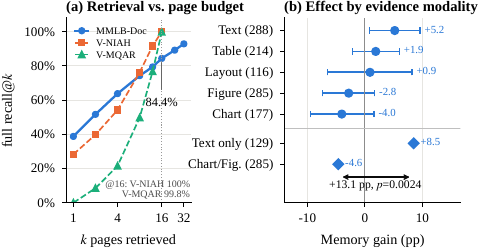}
\caption{Retrieval and modality. \textbf{(a)} ColQwen2 full recall@$k$: V-NIAH and V-MQAR
($1173$ each), and MMLongBench-Doc ($845$ questions with gold pages).
\textbf{(b)} MMLongBench-Doc memory $-$ images by evidence
tag (tags overlap, $95\%$ CIs) and, below the rule, on the $414$ single-tag questions: text
against chart/figure differs by $+13.1$ pp.}
\label{fig:external}
\end{figure}

MMLongBench-Doc differs from FoveDoc-Bench in the two respects that matter here: its evidence is
mostly graphical, and its retrieval is not saturated: full recall@16 is $0.844$
(Fig.~\ref{fig:external}a) and falls with document length, from $0.94$ on documents of at most
$30$ pages to $0.64$ above $120$. Of the $1091$ questions, $244$ are deliberately unanswerable, so
scoring them measures abstention rather than reading, and $246$ carry an empty gold-evidence
set; the reading stratum is the $838$ questions that are both answerable and annotated, each
tagged by the benchmark with the modality of its evidence.

\noindent\textbf{The benefit follows the evidence modality.} On the $414$ questions carrying
exactly one tag (Fig.~\ref{fig:external}b), the paired effect is $+0.085$ on text ($n{=}129$)
against $-0.046$ on chart or figure ($n{=}285$), a difference of $\mathbf{+0.131}$
$[+0.046,+0.215]$ ($\mathbf{p{=}0.0024}$). The benchmark's own overlapping tags order the same
way: plain text
$+0.052$, table $+0.019$, figure $-0.028$, chart $-0.040$. Text evidence gains, as on
FoveDoc-Bench; a table, which is text in a grid, sits between; graphical evidence loses, since a
flat transcript discards the structure a chart or figure encodes~\cite{chartqa,twint}.

\noindent\textbf{The two effects cancel in aggregate.} Chart and figure evidence weighs $55\%$ of
the reading questions against $23\%$ text-only, so the negative stratum carries over twice the
weight of the positive one and the pooled advantage is $+0.004$ (Table~\ref{tab:external}).
The aggregate reports the evidence mix: extracted text is the right
representation for textual evidence, and graphical evidence needs one that keeps its structure.

\noindent\textbf{Text also shifts abstention.} On the unanswerable stratum the memory arm leads
by $+0.049$ ($p{=}0.014$, above the corrected $\alpha$): where the image arm confabulates a
number, the memory arm more often reports that the text contains none. Its abstention hit rate rises from $0.029$ to $0.086$ while
its false-alarm rate moves from $0.001$ to $0.008$, discrimination rather than a bias shift.

\noindent\textbf{Takeaway.} Extracted text helps textual evidence and hurts graphical evidence;
the $13.1$-point contrast marks a modality boundary that an aggregate over a mixed benchmark
hides.

\section{The retrieval boundary}
\label{sec:recall}

\begin{figure*}[t]
\centering
\includegraphics[width=.96\textwidth]{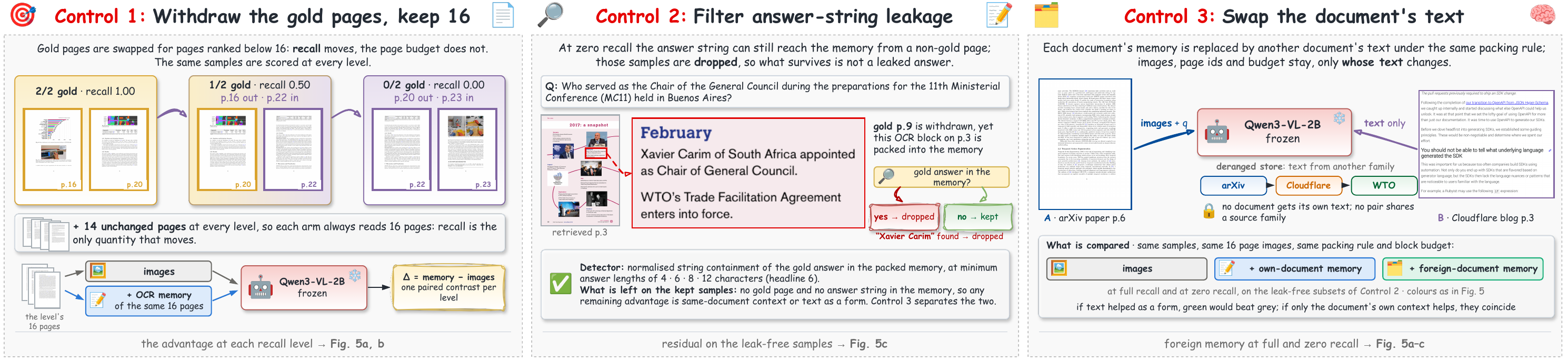}
\caption{Three controls on the memory advantage, all with OCR memory, each on a real sample;
results in Fig.~\ref{fig:sweep}. \textbf{Control 1} varies recall at a fixed $16$-page
budget on the same samples. \textbf{Control 2} drops zero-recall samples whose packed memory
contains the answer string. \textbf{Control 3} builds the memory from another document's text
under the same rule.}
\label{fig:controls}
\end{figure*}

Table~\ref{tab:external} carries a second signal: in the partial-recall stratum the memory arm
trails the image arm by $0.061$, against a lead of $+0.016$ where recall is complete. Packed text
about the wrong pages misleads more than a page image the reader can decline to attend to, cf.\
distractor sensitivity in long contexts~\cite{lostmiddle,distract}, so retrieval quality should
govern the advantage. FoveDoc-Bench's annotations let us measure how directly, by degrading
retrieval under control (Fig.~\ref{fig:controls}).

\begin{figure*}[t]
\centering
\includegraphics[width=\textwidth]{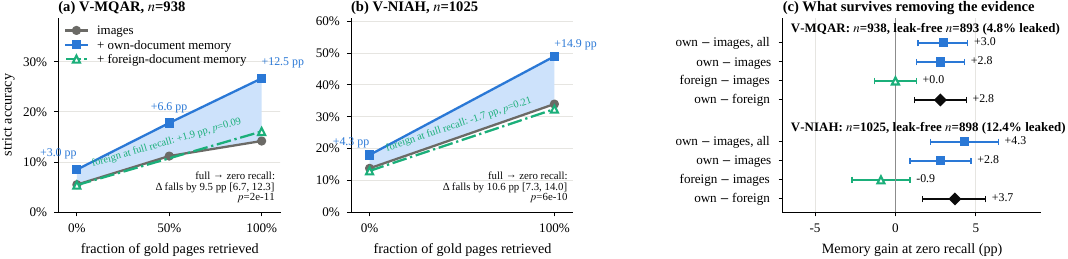}
\caption{Results of the controls in Fig.~\ref{fig:controls}. \textbf{(a, b)} The memory
advantage (labelled) falls monotonically as gold pages are withdrawn; foreign-document text
tracks the image arm at both endpoints. \textbf{(c)} At zero recall the leak-free residual
survives, but foreign-document memory of identical form is indistinguishable from the image
arm: what remains is same-document context, not format. Error bars are $95\%$ CIs.}
\label{fig:sweep}
\end{figure*}

\noindent\textbf{Design.} Retrieval on FoveDoc-Bench is saturated, so we degrade it by hand.
\textbf{Control 1} withdraws gold pages from the retrieved set and substitutes pages the
retriever ranked below the top $16$, holding the page budget at $16$ so that recall is the only
quantity that moves, whereas varying $k$ would confound recall with context length. It scores
the same samples at every level, so the pairing is exact twice over, across arms within a level
and across levels within a sample, and the change in the advantage between full and zero recall
is a within-sample interaction: the part of the advantage that the retrieved gold pages buy.
\textbf{Control 2} restricts zero recall to samples whose packed memory does not contain the
answer string, detected by normalised string containment at four minimum answer lengths, since
values recur in abstracts, headers and cross-references, so that the residual is not a leaked
answer. \textbf{Control 3} replaces each document's memory with a \emph{different} document's
text under the same packing rule and budget; the assignment is deranged, no document receiving
its own text and no pair sharing a source family, so foreign memory can neither leak the answer
nor be topically related, and it isolates what a text memory is worth for its form alone.
Eligible samples, whose documents can supply substitutes at every level, number $938$ for
V-MQAR (two gold pages, three levels) and $1025$ for V-NIAH (one gold page, two levels).

\noindent\textbf{Retrieved evidence buys most of the advantage.} The advantage falls monotonically
as gold pages are withdrawn (Fig.~\ref{fig:sweep}a,b), from $+0.125$ through $+0.066$ at half
recall to $+0.030$ on V-MQAR and from $+0.149$ to $+0.043$ on V-NIAH. The within-sample change is
$+0.095$ $[+0.067,+0.123]$ and $+0.106$ $[+0.073,+0.140]$, both $p<10^{-9}$: roughly three
quarters of the full-recall advantage is bought by the gold pages being present. The residual
at zero recall is small but not zero, and part of it is leakage: the gold answer string sits in
the packed zero-recall memory of
$4.8\%$ of V-MQAR samples and $12.4\%$ of V-NIAH samples. On the leak-free subsets ($n{=}893$
and $898$) the residual is $+0.028$ on both tasks ($p{=}6\times10^{-4}$ and $6\times10^{-3}$),
and it stays between $+0.025$ and $+0.031$ as the detector's minimum answer length varies from
$4$ to $12$ characters.

\noindent\textbf{The residual is context, not format.} Foreign memory at zero recall is
indistinguishable from the image arm on both tasks, $+0.000$ $[-0.013,+0.013]$ ($p{=}0.87$) and
$-0.009$ $[-0.027,+0.009]$ ($p{=}0.40$), and own $-$ foreign is the whole of the residual,
$+0.028$ $[+0.012,+0.044]$ and $+0.037$ $[+0.017,+0.056]$, both $p<10^{-3}$: what survives is
the document's own non-gold text.

The control also speaks to Sec.~\ref{sec:boundary}. Even at \emph{full} recall foreign text is
inert rather than harmful ($+0.019$ and $-0.017$, both $p>0.08$): a reader handed a page of
unrelated prose beside the right images simply ignores it. Contrast the chart-and-figure
stratum, where memory of the \emph{correct} pages actively hurt: what damages the reader is
relevant text about the wrong modality, not irrelevant text.

\noindent\textbf{Takeaway.} Three quarters of the advantage is bought by retrieved evidence and
the rest by the document's own context; text as a format, without that content, is worth nothing.

\section{Conclusion}
\label{sec:discussion}

Under a retrieval-controlled paired protocol, vision--language readers leave evidence unused on
pages they already hold. Extracted text recovers much of it: CPU OCR adds $13$ to $16$
points of strict accuracy over the page images alone, an exact text layer roughly doubles that,
and with the text layer the gain appears in all six readers we test across three model
families. The reader can read the evidence it cannot find: crops of it recover most of the text
gain, boxes around it on the page none. Two boundaries govern the benefit. The evidence
must be textual: on charts and figures extracted text is neutral to harmful, and on a mostly
graphical benchmark the two effects cancel. And retrieval must work: three quarters of the
advantage is bought by the gold pages being present, the rest is the document's own context,
and unrelated text of the same form is worth nothing detectable. An extracted-text memory is an amplifier of
retrieval that works, not a substitute for retrieval that does not.

Two practical conclusions follow. An OCR-then-read stage belongs in text-evidenced document QA
over strong retrieval, gated on retrieval quality and evidence modality rather than applied to
every page. And evaluations of such pipelines should be stratified by recall
and modality, since an aggregate over one mix does not transfer: the same intervention measures
$+0.13$ on FoveDoc-Bench and $+0.004$ on MMLongBench-Doc, and both numbers are right.

\clearpage
\section{Compliance with Ethical Standards}
\label{sec:ethics}

This is a computational study on publicly available documents and benchmarks. It involved no human
or animal subjects, and no ethical approval was required.

\bibliographystyle{IEEEbib}
\bibliography{refs}

\end{document}